\pdfoutput=1
\documentclass[11pt]{article}
\usepackage{subfig}
\usepackage{amsmath,amssymb,amsfonts,amsthm}
\usepackage{algorithmic}
\usepackage{algorithm}
\usepackage[margin=1 in]{geometry}
\usepackage{cite}
\usepackage{graphicx}
\usepackage{textcomp}
\usepackage{subfig}
\usepackage{hhline}
\usepackage[usenames,dvipsnames]{color}
\usepackage{diagbox}

\title{Development of Automatic Endotracheal Tube and Carina Detection on Portable Supine Chest Radiographs using Artificial Intelligence}
\author{Chi-Yeh Chen$^\dag$, Min-Hsin Huang$^\ddag$, Yung-Nien Sun$^\dag$, Chao-Han Lai$^\ddag$
\\ $^\dag$ Department of Computer Science and Information
Engineering, \\ National Cheng Kung University, \\
Taiwan, ROC. \\
$^\ddag$ the Department of Surgery, National Cheng Kung University Hospital, College of Medicine, \\ National Cheng Kung University, \\ Taiwan, ROC. \\
chency@csie.ncku.edu.tw, \\icu.tainan@gmail.com, \\ynsun@mail.ncku.edu.tw, \\d303878@mail.hosp.ncku.edu.tw}

\begin{document}




\maketitle
\begin{abstract}
The image quality of portable supine chest radiographs is inherently poor due to low contrast and high noise. The endotracheal intubation detection requires the locations of the endotracheal tube (ETT) tip and carina. The goal is to find the distance between the ETT tip and the carina in chest radiography. To overcome such a problem, we propose a feature extraction method with Mask R-CNN. The Mask R-CNN predicts a tube and a tracheal bifurcation in an image. Then, the feature extraction method is used to find the feature point of the ETT tip and that of the carina. Therefore, the ETT-carina distance can be obtained. In our experiments, our results can exceed 96\% in terms of recall and precision. Moreover, the object error is less than $4.7751\pm 5.3420$ mm, and the ETT-carina distance errors are less than $5.5432\pm 6.3100$ mm. The external validation shows that the proposed method is a high-robustness system. According to the Pearson correlation coefficient, we have a strong correlation between the board-certified intensivists and our result in terms of ETT-carina distance.

\end{abstract}

\section{Introduction}\label{sec:Introduction}
Mechanical ventilation via endotracheal intubation is a life-saving procedure for patients with acute respiratory failure~\cite{pham2017}. However, improper depth of endotracheal tube (ETT) placement is not rare during endotracheal intubation and may cause catastrophic effects if not recognized promptly after intubation~\cite{BRUNEL19891043, BROWN2015363, ono2018}. An ETT being placed in too shallow a position may increase the risk of accidental extubation or air leak. On the contrary, a deeply positioned ETT may lead to a collapse of the nonventilated lung and hyperinflation of the intubated lung with consequent tension pneumothorax. Therefore, it is essential to measure the distance between the ETT tip and the carina on a chest radiograph (CXR) to confirm the ETT is being placed at the proper depth and to avoid potential life-threatening complications in intensive care units (ICUs)~\cite{lotano2000, HOSSEINNEJAD20131181}.

For ICU patients with endotracheal intubation, the CXR is commonly taken by a portable, or bedside, X-ray machine with patients placed in a supine position. Compared to a standard standing CXR, a portable supine CXR usually has technical disadvantages, such as lower contrast and higher noise~\cite{PMID:1901259}. Several anatomical structures, such as the heart, great vessels, and spine, may overlap with the ETT and carina. Also, the existence of medical devices and lines may obscure the position of the ETT tip and carina on portable CXRs. Identifying the positions of the ETT tip and carina can be a laborious task for critical care providers during routine rounds. ETT malposition may be ignored and thus cannot be promptly corrected. This delay may endanger the lives of those who have been critically ill. A computer-aided detection system that could accurately identify the positions of the ETT tip and carina on portable supine CXRs may improve ICU care quality and ensure patient safety~\cite{chen2016, yi2020}. Therefore, the purpose of this study is to develop an AI system to identify the positions of the ETT tip and carina on portable supine CXRs, to measure the distance between the ETT tip and the carina and to recognize improper ETT positioning in the ICU.

Frid-Adar \textit{et al.}~\cite{Frid-Adar2019} generated synthetic ET tubes for adult X-ray images and then used the U-Net to segment the ETT. Lakhani \textit{et al.}~\cite{Lakhani2020} placed chest radiographs into 12 categories, including bronchial insertion and distance from the carina at 1.0-cm intervals (0.0–0.9 cm, 1.0–1.9 cm, etc.) and greater than 10 cm. They used Inception V3 to classify chest radiographs, which implies predicting the ETT-carina distance interval.

The endotracheal intubation detection can be considered a multi-label mask detection problem. Segmentation methods can be categorized into two classes: segmentation-based methods and detection-based methods. Segmentation-based methods predict the category labels of each pixel in the border region of the image. The most famous segmentation-based method is U-Net~\cite{Ronneberger2015}. Detection-based methods are based on state-of-the-art detectors to get the bounding box of each instance and then predict the mask for each bounding box. The detectors include Faster R-CNN~\cite{Ren2015}, R-FCN~\cite{Dai2016} and Cascade-RCNN~\cite{Cai2018}, etc. Based on Faster R-CNN, He \textit{et al.}~\cite{He2017} proposed Mask R-CNN, in which an instance-level semantic segmentation branch is added. According to position-sensitive scores, Chen \textit{et al.}~\cite{Chen2018} proposed MaskLab to obtain better results. However, the score of the instance mask is from the box-level classification confidence. To focus on scoring the masks, Huang \textit{et al.}~\cite{Huang2019} proposed Mask Scoring R-CNN, which is based on Mask R-CNN.

This paper proposes a feature extraction method with Mask R-CNN. The Mask R-CNN predicts a tube and a tracheal bifurcation in an image. Then, the feature extraction method is used to find the feature point of the tube tip and that of the carina. Compared to \cite{Lakhani2020}, this paper predicts not only the ETT–carina distance but also the mask of the distal ETT end, the mask of the tracheal bifurcation, the feature point of the tube tip, and the feature point of the carina. Moreover, the solution of \cite{Lakhani2020} was trained with category labels, whereas our solution was trained using pixel-level segmentation labels, which can generate a more accurate object location.

The rest of this paper is organized as follows. Section \ref{sec:Method} proposes a method for endotracheal intubation detection. Section \ref{sec:Experiments} shows the experiment results. Section \ref{sec:Conclusion} draws conclusions.

\begin{figure}[t]
    \centering
        \includegraphics[width=3 in]{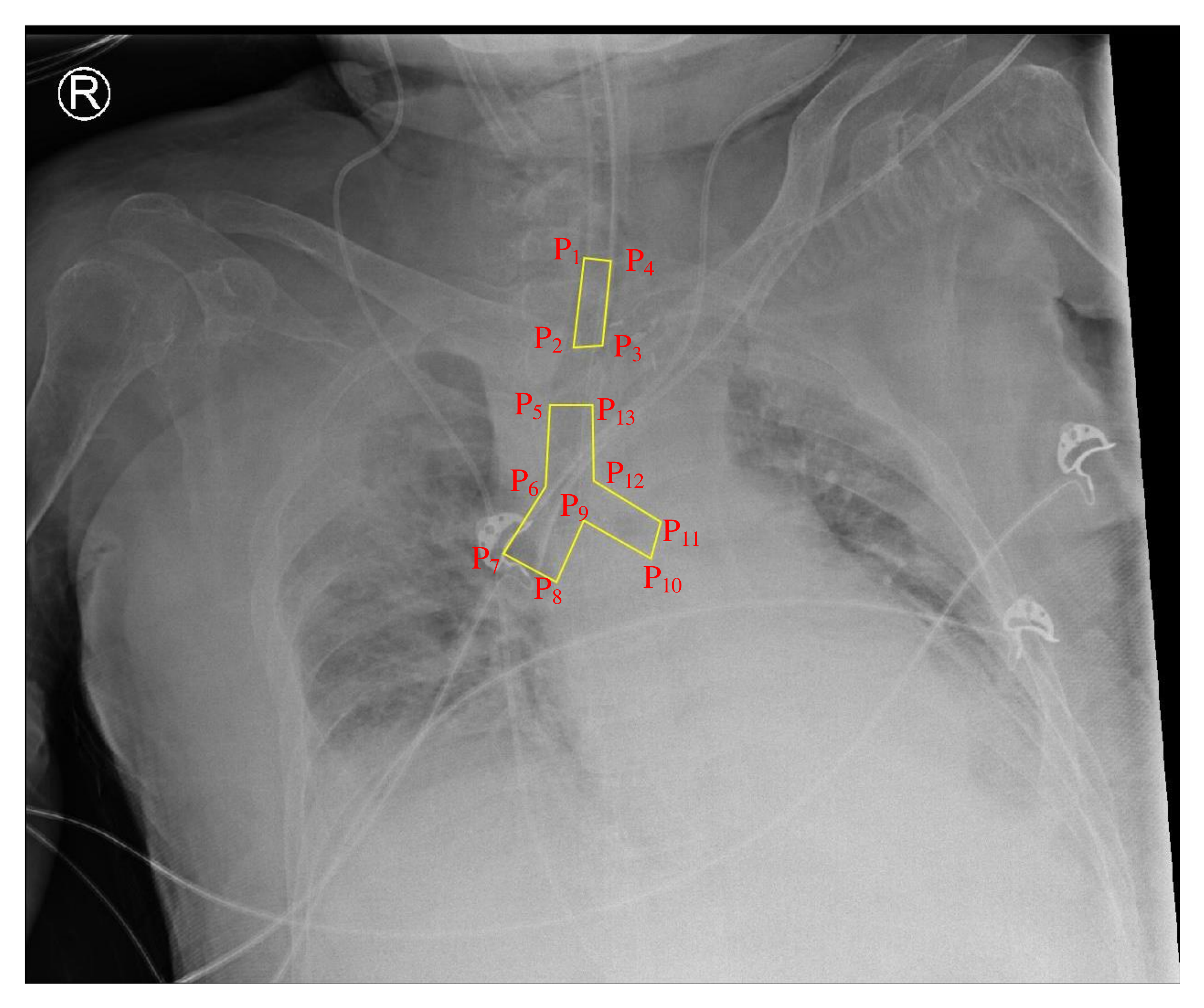} 
        \caption{The ground truth.}
    \label{fig1}
\end{figure}

\begin{figure*}[t]
    \centering
        \includegraphics[width=6 in]{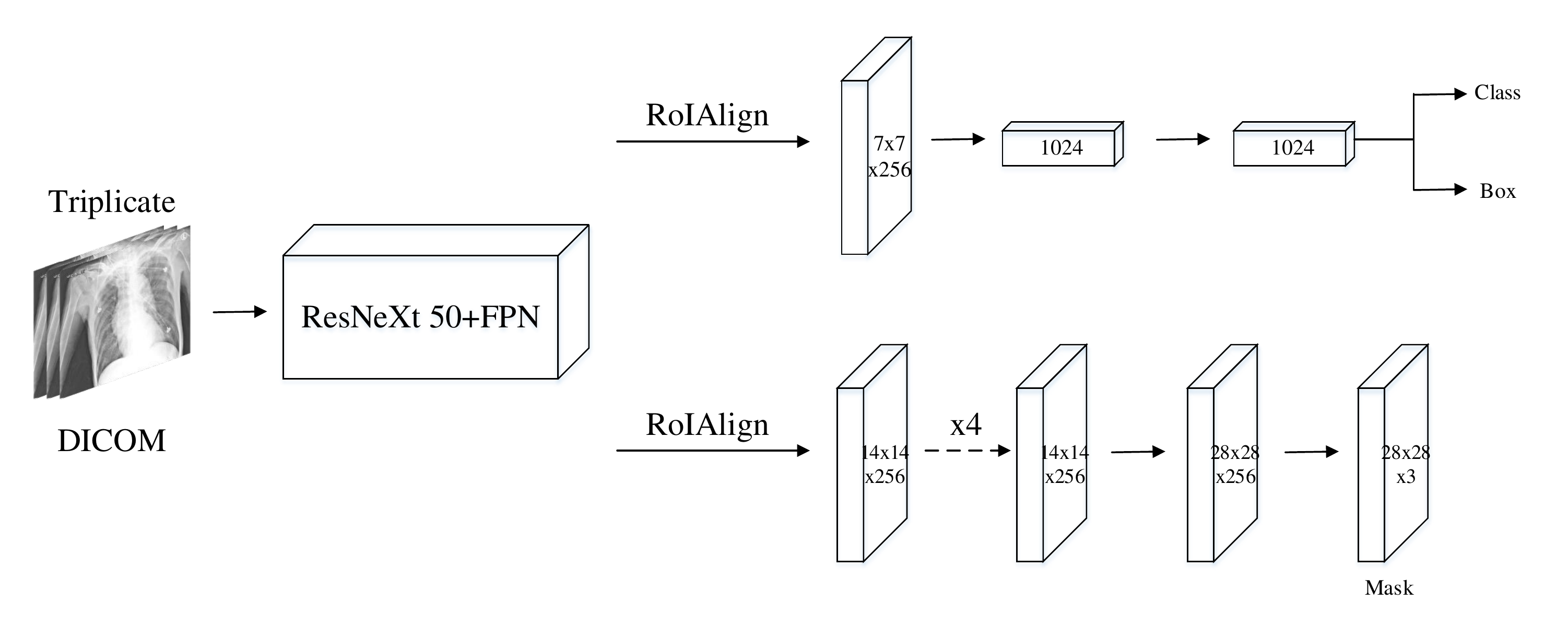} 
        \caption{Mask R-CNN.}
    \label{fig2}
\end{figure*}

\section{Method}\label{sec:Method}
In the ground truth, four points ($P_{1-4}$) are used to label the end part of the ETT, and another nine points ($P_{5-13}$) are used to label the tracheal bifurcation (see Fig.~\ref{fig1}). Moreover, two boxes with a size of $48\times 48$ are used to label the feature point of the ETT tip (the middle point of ($P_2$, $P_3$)) and the feature point of the carina ($P_9$), respectively. The goal is to find the mask of the distal ETT end, the mask of the tracheal bifurcation, and two boxes of the feature point on chest radiographs to locate the ETT tip and the carina.   

This work uses Mask R-CNN~\cite{He2017} to predict the masks of ETT and tracheal bifurcation. Figure~\ref{fig2} shows the Mask R-CNN architecture. The backbone is ResNeXt 50 (32x4d)~\cite{Xie2017} with Feature Pyramid Network (FPN)~\cite{Lin2017}. The head architecture (right panel in Fig.~\ref{fig2}) extended the Faster R-CNN head. The numbers denote the spatial resolution and channels. According to the spatial dimension, arrows are either conv or deconv layers. The single number denotes the fc layer. All convs are $3\times 3$, except the output conv is $1\times 1$. The deconvs are $2\times 2$ with stride 2.

\begin{figure*}[!t]
    \centering
        \includegraphics[width=6 in]{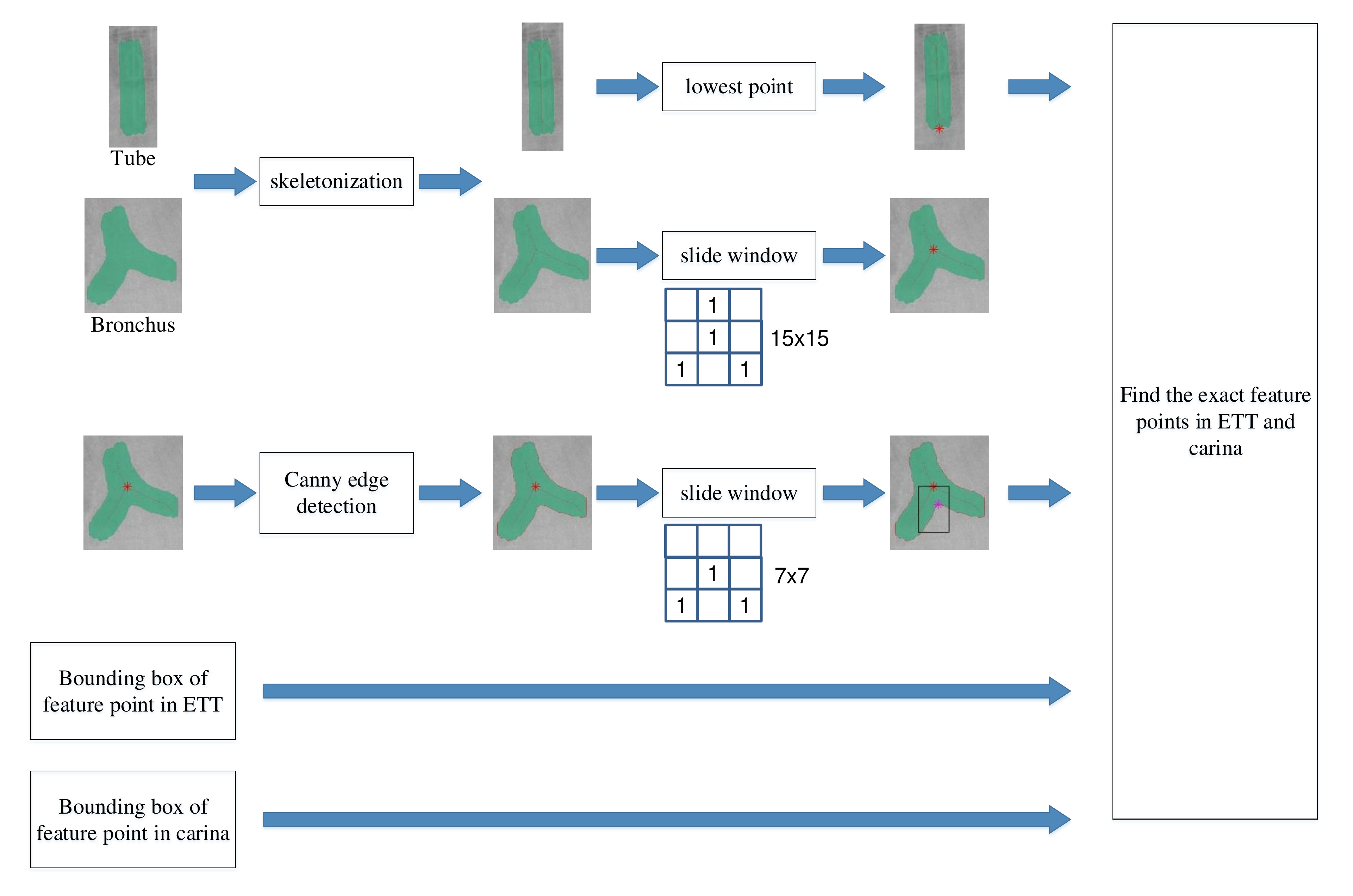} 
        \caption{The processing of finding feature points in the ETT tip and carina.}
    \label{fig3}
\end{figure*}

\subsection{The Inference Process}
In the inference step, we only keep the object with the maximal score for each class. Then, a feature extraction method is used to find the feature points. Figure~\ref{fig3} shows the process of finding the feature points of the ETT tip and carina. We use the boxes of the feature points and the masks, which are predicted by the Mask R-CNN, to obtain the exact feature points of the ETT tip and carina. Two results are obtained by the bounding boxes of the feature point and by the masks of the ETT and the carina, respectively. Then, the exact feature points of the ETT tip and carina can be obtained by fusing the two results.

\begin{figure*}[!t]
    \centering
        \includegraphics[width=3.2 in]{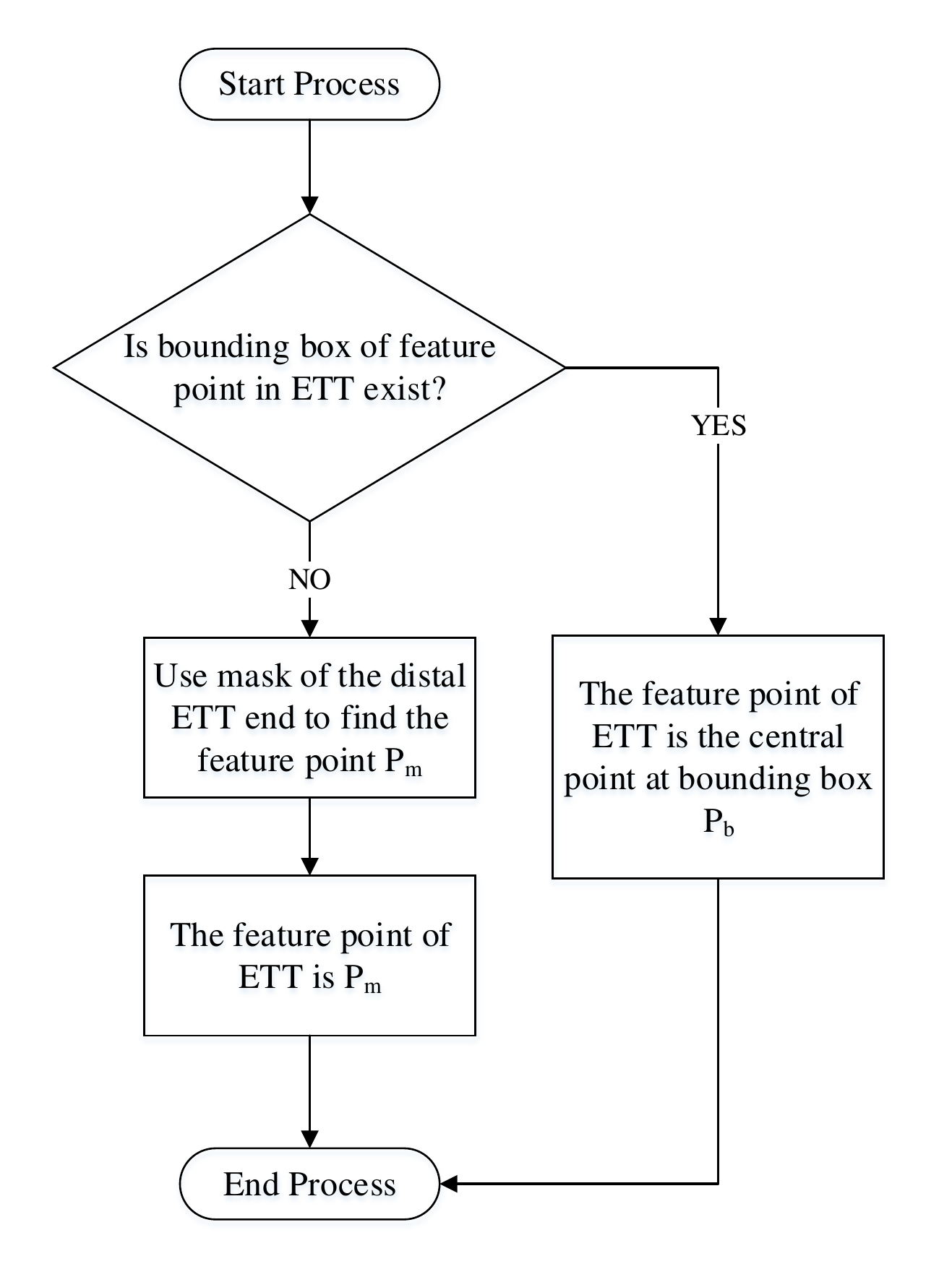} 
        \caption{The flow chart of finding the exact feature points in ETT tip.}
    \label{fig4_1}
\end{figure*}  

\begin{figure*}[!t]
    \centering
        \includegraphics[width=3.6 in]{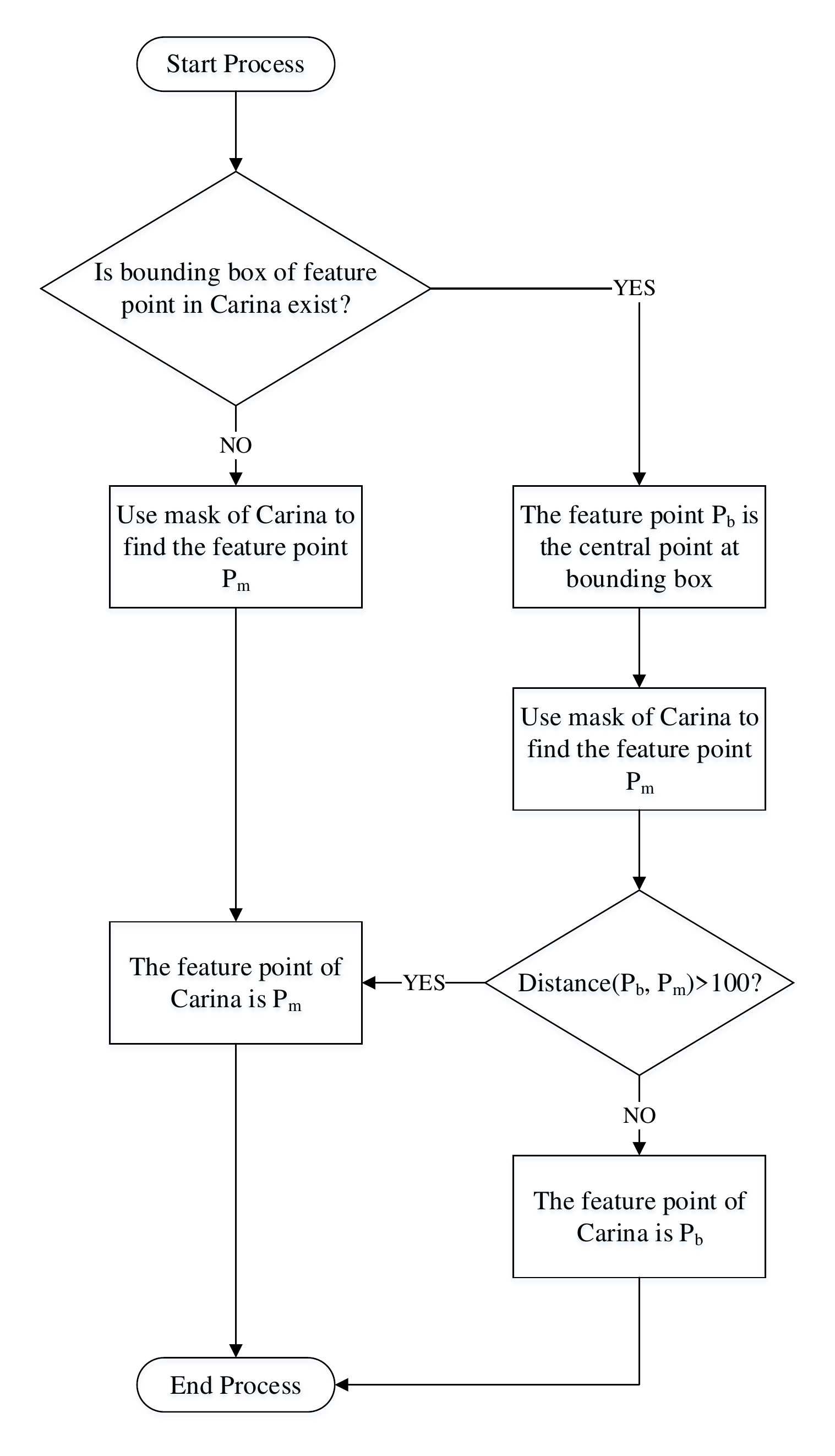} 
        \caption{The flow chart of finding the exact feature points in carina.}
    \label{fig4_2}
\end{figure*}  

The masking processes are as follows. We use skeletonization to find the skeleton of the mask. The feature point of the ETT is the lowest point in the mask skeleton (see the top of Figure~\ref{fig3}). In the feature point of the carina, this work uses a slide window with size of 15x15 to find the central point in the skeleton of the carina mask. Then, the Canny edge detection is applied to find the edge of the mask. According to the central point, a patch of the mask edge with a size of 100x150 is cropped. Then, we used a slide window with a size of 7x7 to find the feature point of the carina in the patch (see the middle of Figure~\ref{fig3}).

Figures~\ref{fig4_1} and~\ref{fig4_2} show the flow chart of finding the exact feature points in the ETT and carina. In the feature points of the ETT tip, we accept all results of the bounding boxes of the feature point. Since the bounding boxes of the feature point are too small, many objects (about 10\%) are not detected. Hence, this work uses the mask result to supplement the bounding box result. In the feature points of the carina, the mean error of the bounding box result is low and the standard deviation is high. On the other hand, the mean error of the mask result is high and standard deviation is low. Hence, we use the mask result to eliminate the worst cases of the bounding box result. If the distance between two feature points (from bounding box and mask) is greater than 100 pixels, we use the mask result to replace the bounding box result. Moreover, the mask result also supplements the bounding box result.

\section{Experiments}\label{sec:Experiments}

\subsection{Dataset and Evaluation Metrics}\label{sec:Preliminaries}
This study was approved by the institutional review board (IRB) of the National Cheng Kung University Hospital (IRB number: A-ER-108-305). Chest radiographs of intubated ICU patients were identified by data research in the institutional Picture Archiving and Communication System (PACS), and the underlying DICOMs were exported after de-identification. The dataset had a total of 1,842 portable chest radiographs. The ground-truth annotations were labeled by two board-certified intensivists. Moreover, we used 150 images from outsider hospitals to validate.

Four metrics, namely object error, distance error, recall, and precision, were applied to evaluate the performance of the proposed method. The concept of Dice Coefficient (DC) was used to measure the performance of mask (the two polygons representing the end part of the ETT end and the bifurcation of the tracheobronchial tree) detection. The DC can measure the similarity between two sets of data and has been broadly used for validating image segmentation algorithms. Moreover, the Euclidean Distance was used to measure the performance of the feature point (ETT tip and carina) detection.

Let $MP$ be the middle point of $(P_2, P_3)$ in the ground truth annotation. The distance between the point $MP$ and the predicted feature point of the ETT tip is the object error of the ETT tip. Moreover, the distance between the point $P_9$ in the annotated chest radiographs and the predicted feature point of the carina is the object error of the carina. We consider the object to have been successfully detected if the Dice Coefficient was no less than 0.6 or the object error was no more than 100 pixels. We call the detected object according to a true positive. A false positive is an error in object detection in which a mask incorrectly indicates an object. Moreover, a false negative is the error in which the model did not indicate the object. The definitions of recall and precision are:
\begin{eqnarray*}
recall=\frac{TP}{TP+FN}
\end{eqnarray*}
and
\begin{eqnarray*}
precision=\frac{TP}{TP+FP}
\end{eqnarray*}
where $TP$ is the number of true positives, $FN$ is the number of false negatives, and $FP$ is the number of false positives.

Let $d_1$ be the ETT-carina distance between the point $MP$ and the point $P_9$ in the ground truth annotation. Let $d_2$ be the ETT–carina distance between the predicted feature point of the ETT tip and the predicted feature point of the carina. The ETT–carina distance error was the absolute value of $d_1-d_2$.

\subsection{Implementation Details}
The MMDetection~\cite{mmdetection} and the PyTorch library were used for building the network. The layers of the network were initialized using pretrained weights from open-mmlab. The loss functions of the bounding box head were set to the cross-entropy and smooth L1. The loss function of the mask head was set to the average binary cross-entropy. Images were resized to $1258\times 1516$, and each mini batch has 5 images. The model was trained using an SGD optimizer with a momentum of 0.9 and a weight decay of 0.0001. The learning rate was 0.001. Training was performed for 120 epochs, decreasing the learning rate by a factor of 0.1 after 80 epochs and 110 epochs.

\begin{table}[!th]
\caption{The object detection performance in recall and precision}
\centering
\begin{tabular}{|l|c|c|c|c|}
\hline
        & \multicolumn{2}{c|}{Tube Tip} & \multicolumn{2}{c|}{Carina} \\ \hline
        &    Recall          &    Precision             &  Recall         &   Precision        \\ \hline
 Fold 1 &    95.42\%	& 95.53\% & 95.74\%	& 96.28\% \\ \hline
 Fold 2 &    95.84\%	& 96.26\% & 96.86\%	& 97.52\% \\ \hline
 Fold 3 &    96.26\%	& 96.47\% & 96.71\%	& 97.26\% \\ \hline
 Fold 4 &    96.96\%	& 96.96\% & 97.10\%	& 97.54\% \\ \hline
 Average &    96.12\%	& 96.31\% & 96.60\%	& 97.15\% \\ \hhline{|=|=|=|=|=|}
 External val. & 93.96\%	& 93.96\% & 95.47\%	& 95.80\% \\ \hline
\end{tabular}
\label{tab:ratio1}
\end{table}

\subsection{Results}
Table~\ref{tab:ratio1} shows the object detection performance in terms of recall and precision. The recall of the ETT tip was 96.12\%, and the precision of that is 96.31\%. The recall of the carina was 96.60\%, and the precision of that was 97.15\%. In an external validation, the performance degraded less than 3\%

\begin{table}[!th]
\caption{The object detection performance in object error}
\centering
\begin{tabular}{|l|c|c|c|c|}
\hline
        & \multicolumn{2}{c|}{Tube Tip (mm)} & \multicolumn{2}{c|}{Carina  (mm)} \\ \hline
        &    Mean            &    Std.                  &  Mean           &   Std.        \\ \hline
 Fold 1 &    4.1219	& 4.6022 & 4.9794 & 5.3327          \\ \hline
 Fold 2 &    4.2362	& 4.7496 & 4.5163	& 5.5910           \\ \hline
 Fold 3 &    4.2715	& 4.5649 & 4.8013	& 5.4726           \\ \hline
 Fold 4 &    3.8564	& 3.6895 & 4.8032	& 4.9717           \\ \hline
 Average&    4.1215	& 4.4016 & 4.7751	& 5.3420           \\ \hhline{|=|=|=|=|=|}
 External val. & 4.2856	& 5.9425 & 4.5668 & 4.5132 \\ \hline
\end{tabular}
\label{tab:ratio2}
\end{table}

Table~\ref{tab:ratio2} shows the object detection performance in terms of object error. In the ETT tip location, the mean of object error was 4.1215 mm, and the standard deviation of that was 4.4016 mm. In the carina location, the mean of the object error was 4.7751 mm, and the standard deviation of that was 5.3420 mm. In an external validation, the performance of the ETT tip location decreased slightly. Table~\ref{tab:ratio3} shows the object detection performance in terms of the ETT–carina distance error. The mean of the ETT–carina distance error was 5.5432 mm. The standard deviation of the ETT–carina distance error was 6.3100 mm. In an external validation, the performance also decreased slightly. In \cite{Lakhani2020}, the mean of the ETT–carina distance error was 6.9 mm, and the standard deviation of the ETT–carina distance error was 7.0 mm.

\begin{table}[!th]
\caption{The object detection performance in ETT-carina distance error}
\centering
\begin{tabular}{|l|c|c|}
\hline
        &    Mean (mm)  &    Std. (mm) \\ \hline
 Fold 1 &    5.7413	& 6.3497  \\ \hline
 Fold 2 &    5.6370	& 7.3877  \\ \hline
 Fold 3 &    5.5754	& 6.3576  \\ \hline
 Fold 4 &    5.2190	& 5.1451  \\ \hline
 Average&    5.5432	& 6.3100  \\ \hhline{|=|=|=|}
 External val. & 5.6680	& 6.6514 \\ \hline
\end{tabular}
\label{tab:ratio3}
\end{table}

\begin{table}[!th]
\caption{The distribution of number of images in ETT-carina distance error}
\centering
\begin{tabular}{|l|c|c|c|c|}
\hline
        &    $\leq 5$ mm & $\leq 10$ mm &  $\leq 15$ mm &   $\leq 20$ mm        \\ \hline
 Fold 1 &    59.31\%	& 84.42\%	& 91.13\%	& 95.24\%          \\ \hline
 Fold 2 &    60.30\%	& 83.30\%	& 92.62\%	& 95.44\%           \\ \hline
 Fold 3 &    59.83\%	& 83.41\%	& 92.58\%	& 94.76\%           \\ \hline
 Fold 4 &    62.04\%	& 85.68\%	& 94.79\%	& 96.10\%           \\ \hline
 Average&    60.37\%	& 84.20\%	& 92.78\%	& 95.39\%           \\ \hhline{|=|=|=|=|=|}
 External val. & 64.00\%	& 82.00\%	& 90.67\%	& 94.67\% \\ \hline
\end{tabular}
\label{tab:ratio4}
\end{table}

Table~\ref{tab:ratio4} shows the distribution of the number of images in terms of ETT–carina distance error. We have 92.78\% images that are less than 15 mm in ETT–carina distance error. Table~\ref{tab:ratio5} shows the distribution of the number of images in object error (Endotracheal tube tip). We have 96.36\% images that are less than 15 mm in object error. Table~\ref{tab:ratio6} shows the distribution of the number of images in object error (carina). We have 95.55\% images that are less than 15 mm in object error. Moreover, an external validation shows that the performance has no significant differences.

\begin{table}[!th]
\caption{The distribution of number of images in object error (Endotracheal tube tip)}
\centering
\begin{tabular}{|l|c|c|c|c|}
\hline
        &    $\leq 5$ mm & $\leq 10$ mm &  $\leq 15$ mm &   $\leq 20$ mm        \\ \hline
 Fold 1 &    77.06\%	& 92.86\%	& 96.54\%	& 97.62\%          \\ \hline
 Fold 2 &    73.75\%	& 91.76\%	& 95.66\%	& 97.61\%           \\ \hline
 Fold 3 &    72.71\%	& 92.36\%	& 95.85\%	& 98.03\%          \\ \hline
 Fold 4 &    76.79\%	& 92.84\%	& 97.40\%	& 99.57\%           \\ \hline
 Average&    75.08\%	& 92.46\%	& 96.36\%	& 98.21\%           \\ \hhline{|=|=|=|=|=|}
 External val. & 79.33\%	& 90.00\%	& 95.33\%	& 96.67\%  \\ \hline
\end{tabular}
\label{tab:ratio5}
\end{table}

\begin{table}[!th]
\caption{The distribution of number of images in object error (carina)}
\centering
\begin{tabular}{|l|c|c|c|c|}
\hline
        &    $\leq 5$ mm & $\leq 10$ mm &  $\leq 15$ mm &   $\leq 20$ mm        \\ \hline
 Fold 1 &    67.53\%	& 90.04\%	& 95.02\%	& 96.54\%          \\ \hline
 Fold 2 &    70.93\%	& 93.06\%	& 96.10\%	& 97.83\%           \\ \hline
 Fold 3 &    69.65\%	& 91.27\%	& 94.76\%	& 96.72\%          \\ \hline
 Fold 4 &    67.25\%	& 91.76\%	& 96.31\%	& 97.40\%           \\ \hline
 Average&    68.84\%	& 91.53\%	& 95.55\%	& 97.12\%           \\ \hhline{|=|=|=|=|=|}
 External val. & 73.33\%	& 89.33\%	& 95.33\%	& 96.54\%  \\ \hline
\end{tabular}
\label{tab:ratio6}
\end{table}

\begin{table}[!th]
\caption{The confusion matrix of diagnosis}
\centering
\begin{tabular}{|l|c|c|}
\hline
 \diagbox{Predict}{GT} & Suitable &  Unsuitable \\ \hline
 Suitable   &  1305 &	107  \\ \hline
 Unsuitable &  85  &	318  \\ \hline
 Undetection&  12	  & 15   \\ \hline
\end{tabular}
\label{tab:ratio7}
\end{table}

Table~\ref{tab:ratio7} shows the confusion matrix of diagnosis. We call an ETT suitable if the ETT–carina distance is in the range between 20 to 70 mm. We have 1,623 (88.11\%) images that correspond to results from the board-certified intensivists. Table~\ref{tab:ratio8} shows that the correlation between the ground truth and the result of the proposed method is significant in terms of ETT–carina distance. In the external validation, Table~\ref{tab:ratio9} also shows that the correlation between the ground truth and the result of the proposed method is significant in terms of ETT-carina distance.

\begin{table*}[!th]
\caption{The correlation between ground truth and the result of proposed method in terms of ETT-carina distance.}
\centering
\begin{tabular}{|r|c|c|c|c|}
\hline
                            & fold 1           & fold 2           & fold 3           & fold 4           \\ \hline
\multicolumn{5}{|l|}{\textbf{Pearson $r$}}                                                                \\ \hline
$r$                           & 0.8895           & 0.8742           & 0.8800           & 0.9164           \\ \hline
95\% confidence interval    & 0.8686 to 0.9072 & 0.8505 to 0.8943 & 0.8573 to 0.8993 & 0.9004 to 0.9300 \\ \hline
R square                    & 0.7912           & 0.7642           & 0.7744           & 0.8398           \\ \hline
\multicolumn{5}{|l|}{\textbf{P value}}                                                                  \\ \hline
P (two-tailed)              & \textless 0.0001 & \textless 0.0001 & \textless 0.0001 & \textless 0.0001 \\ \hline
P value summary             & ****             & ****             & ****             & ****             \\ \hline
Significant? ($\alpha=0.05$) & Yes              & Yes              & Yes              & Yes              \\ \hline
\multicolumn{5}{|l|}{}                                                                                  \\ \hline
Number of XY Pairs          & 456              & 451              & 451              & 457              \\ \hline
Overall number              & 462              & 461              & 458              & 460              \\ \hline
\end{tabular}
\label{tab:ratio8}
\end{table*}

\begin{table}[!th]
\caption{The correlation between ground truth of outsider hospitals and the result of proposed method in terms of ETT-carina distance.}
\centering
\begin{tabular}{|r|c|}
\hline
                            & Other Hospital Data \\ \hline
\multicolumn{2}{|l|}{\textbf{Pearson $r$}}       \\ \hline
$r$                           & 0.8860           \\ \hline
95\% confidence interval    & 0.8457 to 0.9163 \\ \hline
R square                    & 0.7851           \\ \hline
\multicolumn{2}{|l|}{\textbf{P value}}         \\ \hline
P (two-tailed)              & \textless 0.0001 \\ \hline
P value summary             & ****             \\ \hline
Significant? ($\alpha=0.05$) & Yes              \\ \hline
\multicolumn{2}{|l|}{}                         \\ \hline
Number of XY Pairs          & 149              \\ \hline
Overall number              & 150              \\ \hline
\end{tabular}
\label{tab:ratio9}
\end{table}

Table~\ref{tab:ratio10} shows the visualization of result. Although the first two images of the good case are blurred, the features of the ETT and the carina are clear. The last image of the good case is an easy case. The images of the medium case were selected based on the mean error. Although the predicted feature points had a slight error in the medium case, the predicted shape of the objects was correct. Due to blurred and unclear features in the first two images of the worst case, the proposed method cannot obtain an exact result. In the last image of the worst case, the predicted ETT was incorrect, since the edge of the end part is unclear.

\begin{table*}[!th]
\caption{The predicted images: the yellow line is ground truth, the yellow asterisk is ground truth of feature point and the red asterisk is predicted feature point.}
\begin{tabular}{|l|c|c|c|}
\hline
 Good   &  \includegraphics[width=2 in]{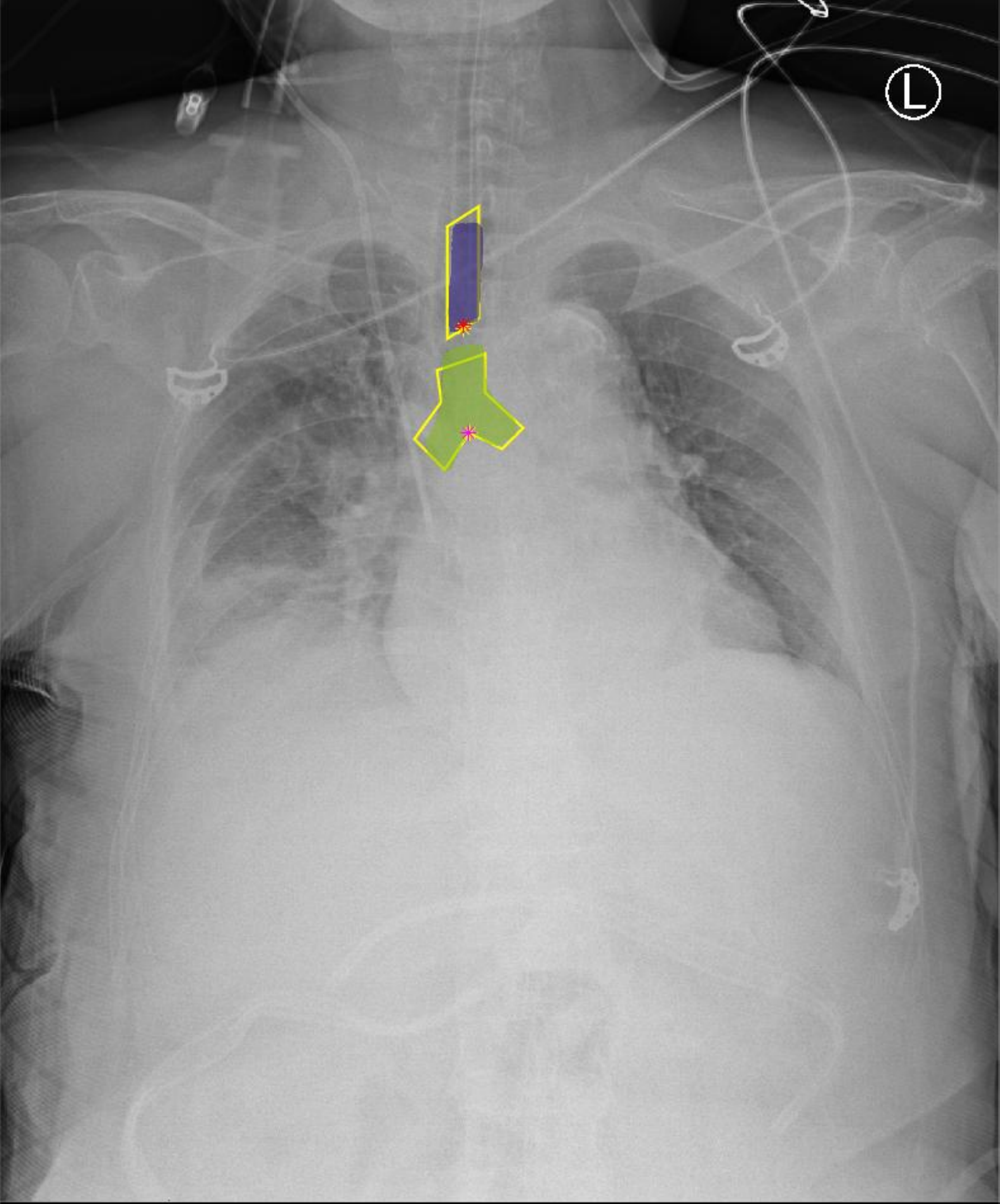}  & \includegraphics[width=2 in]{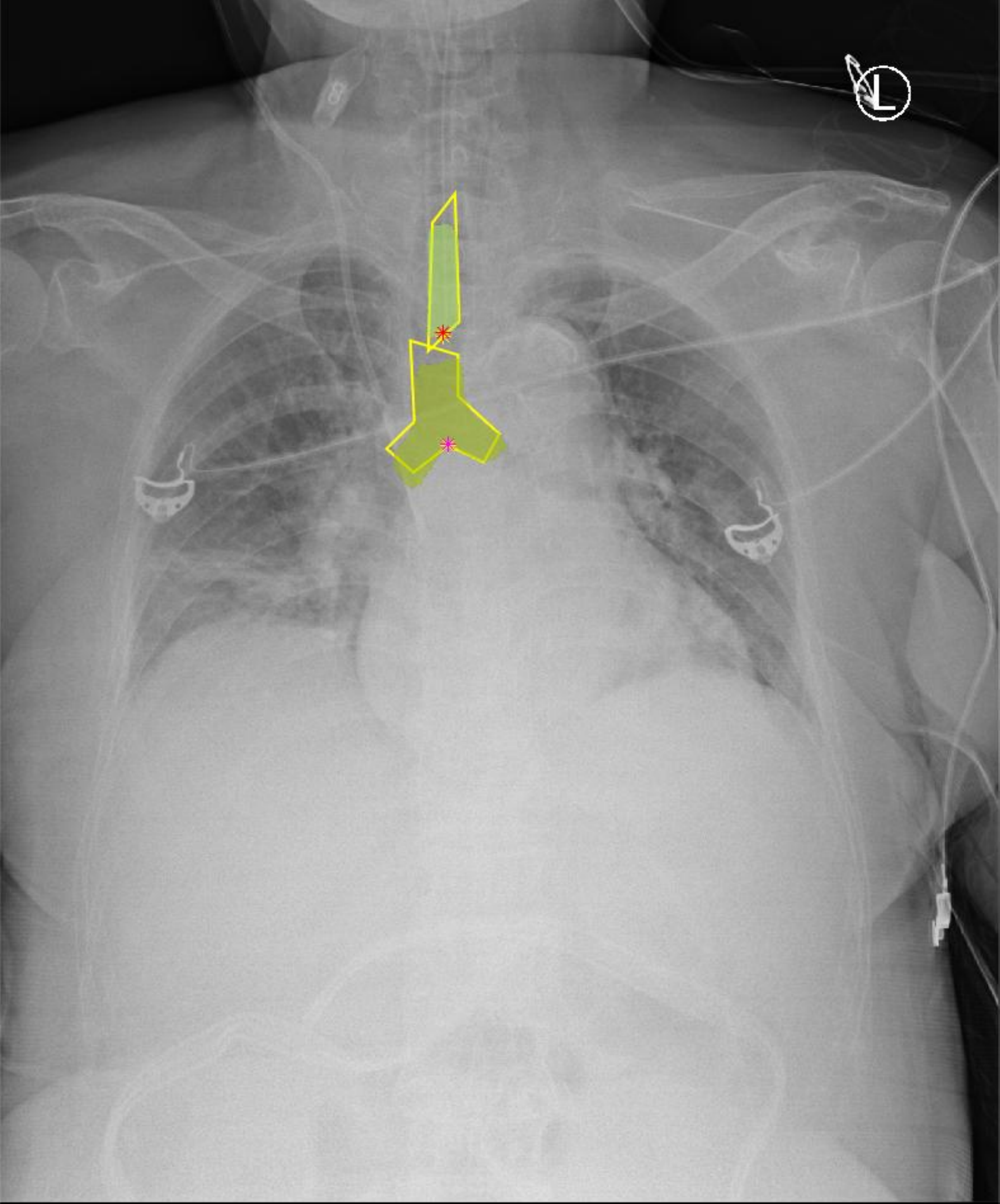} &  \includegraphics[width=2 in]{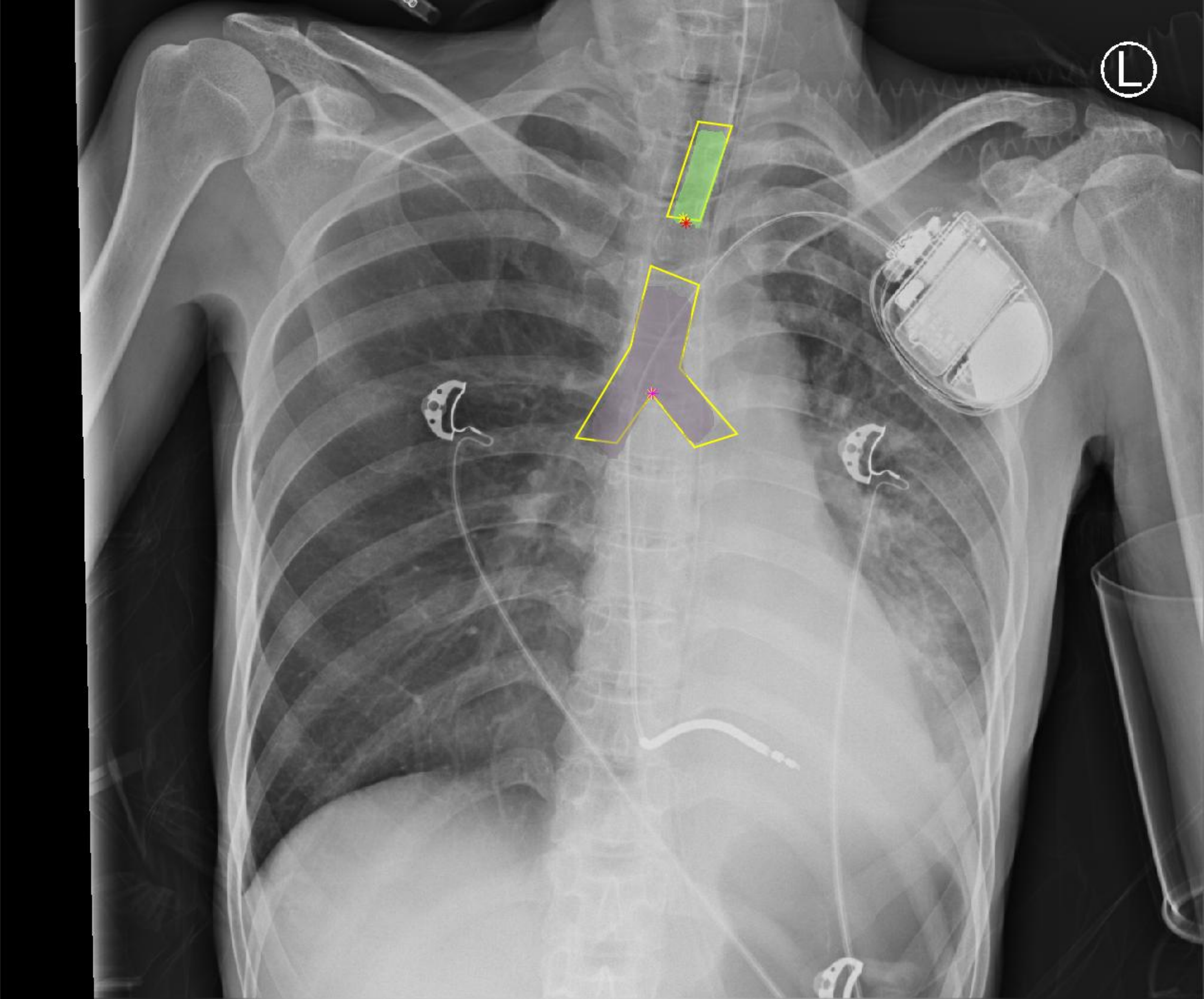}\\ \hline
 Medium & \includegraphics[width=2 in]{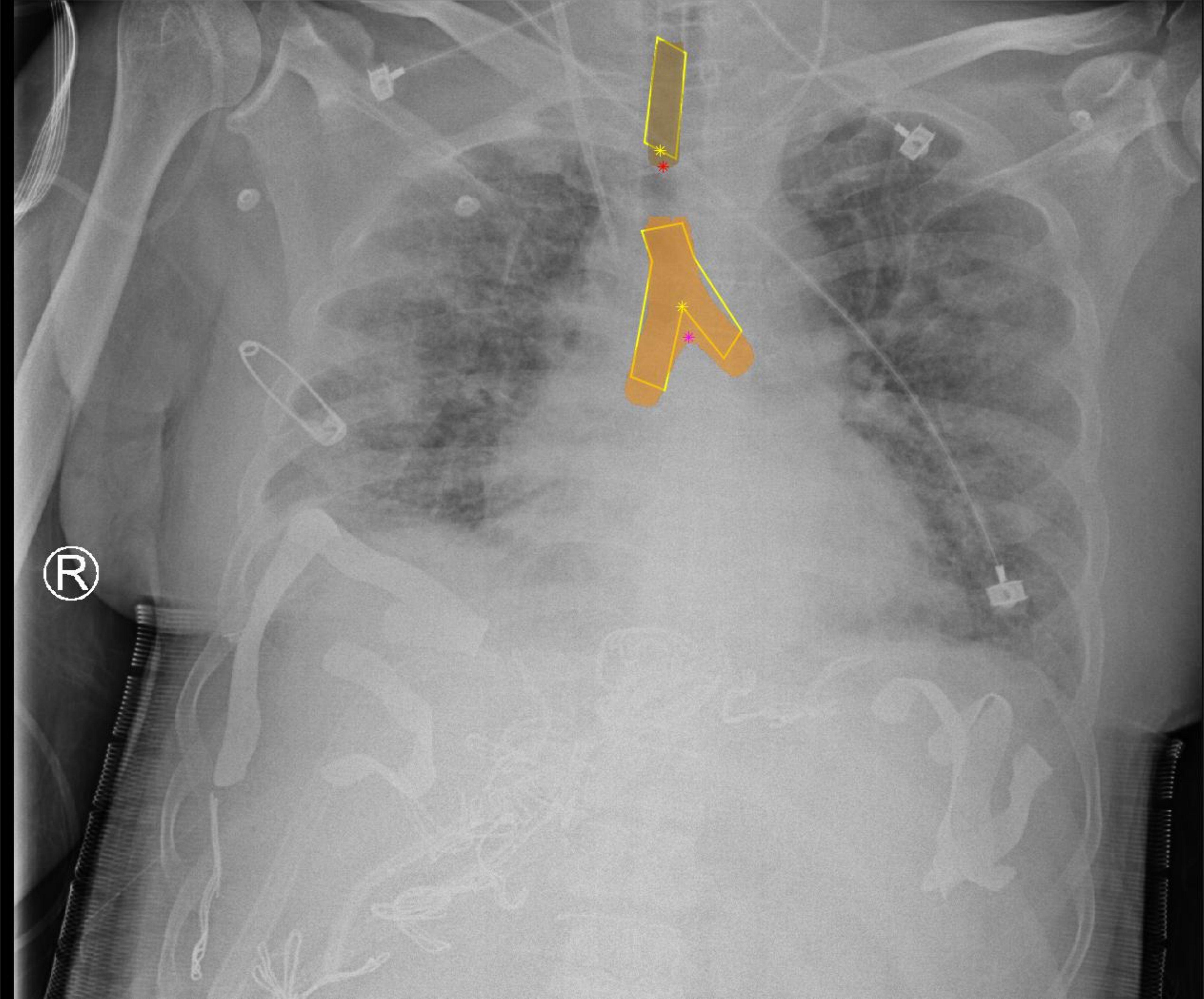} & \includegraphics[width=2 in]{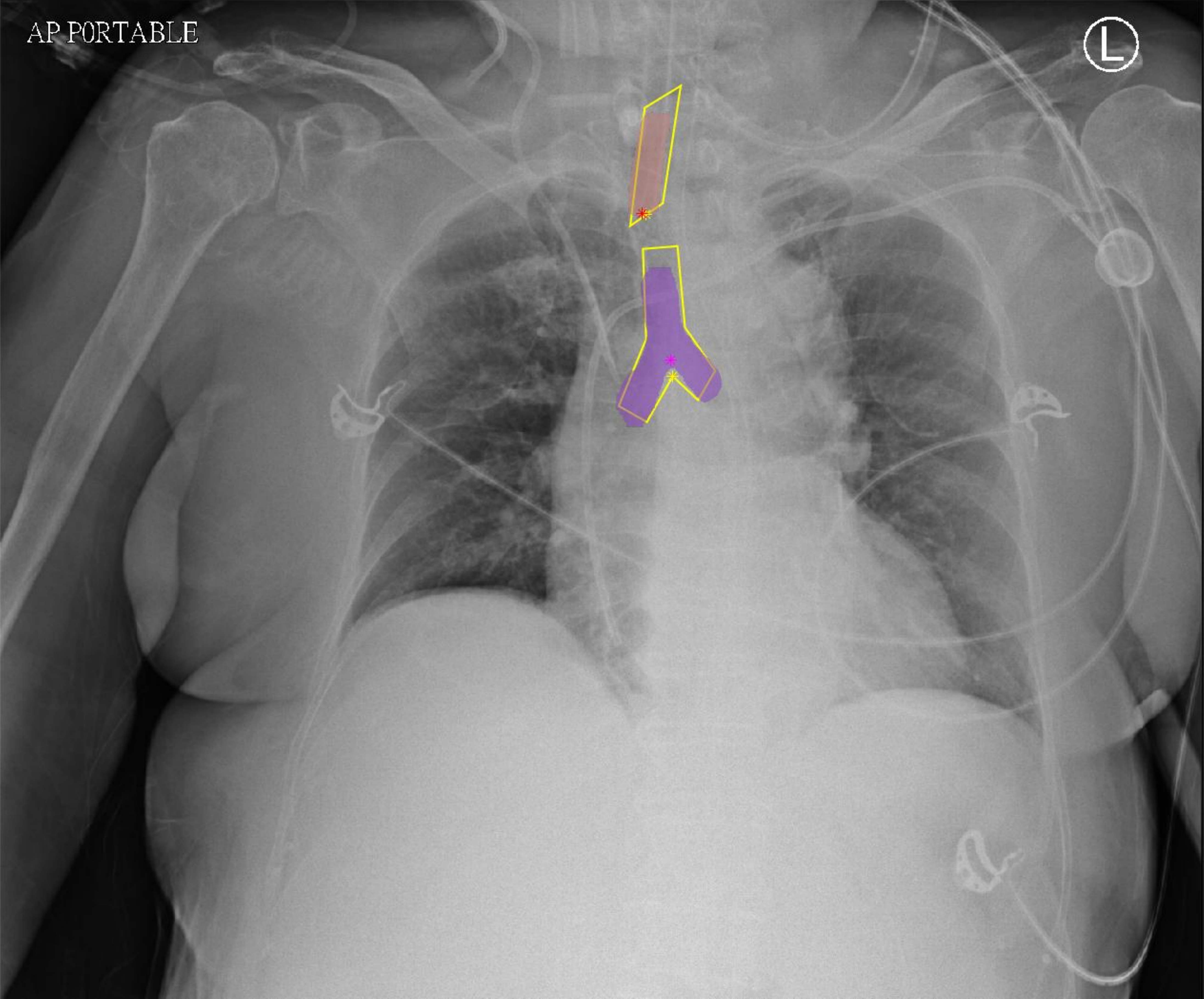} & \includegraphics[width=2 in]{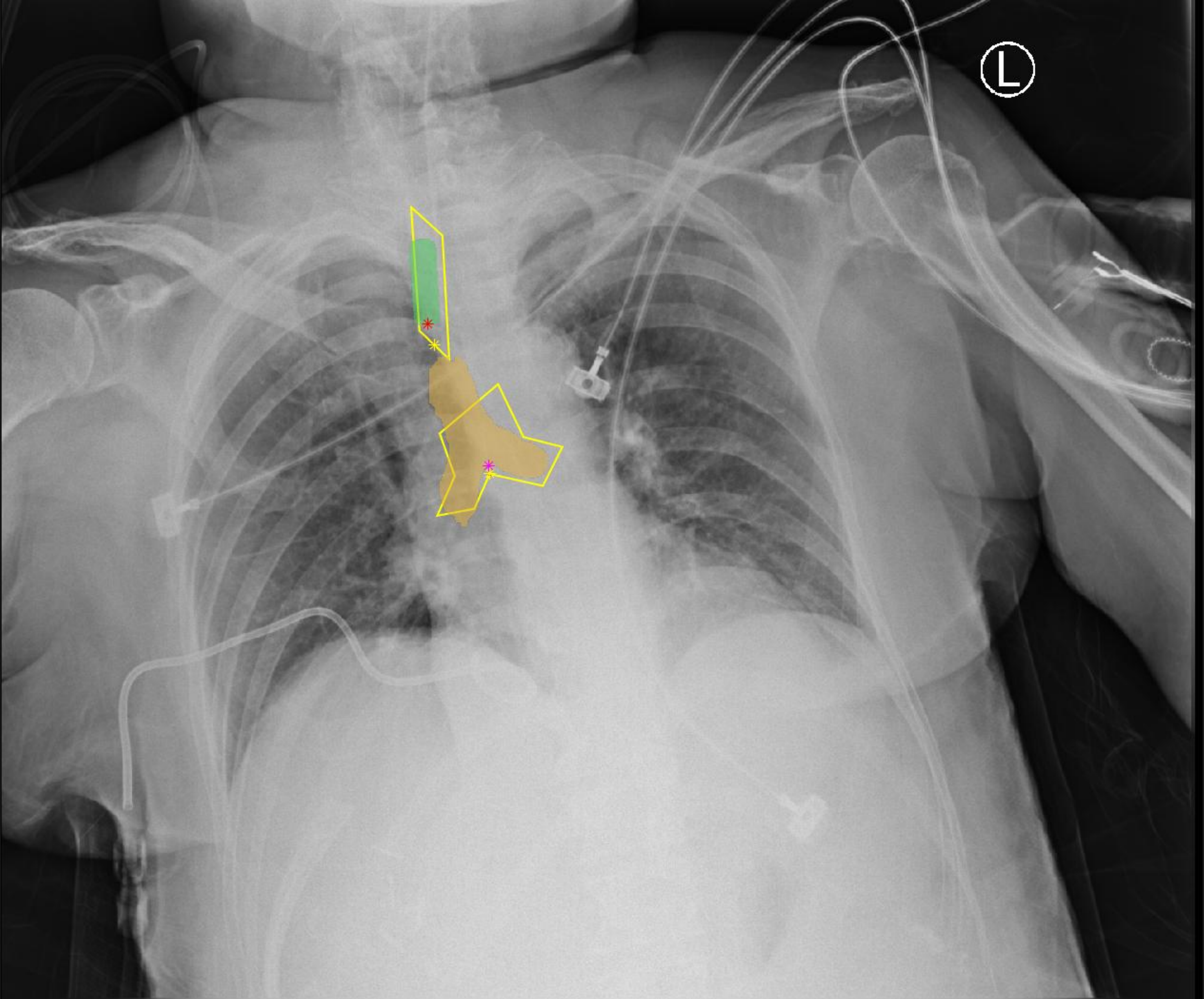} \\ \hline
 Worse  & \includegraphics[width=2 in]{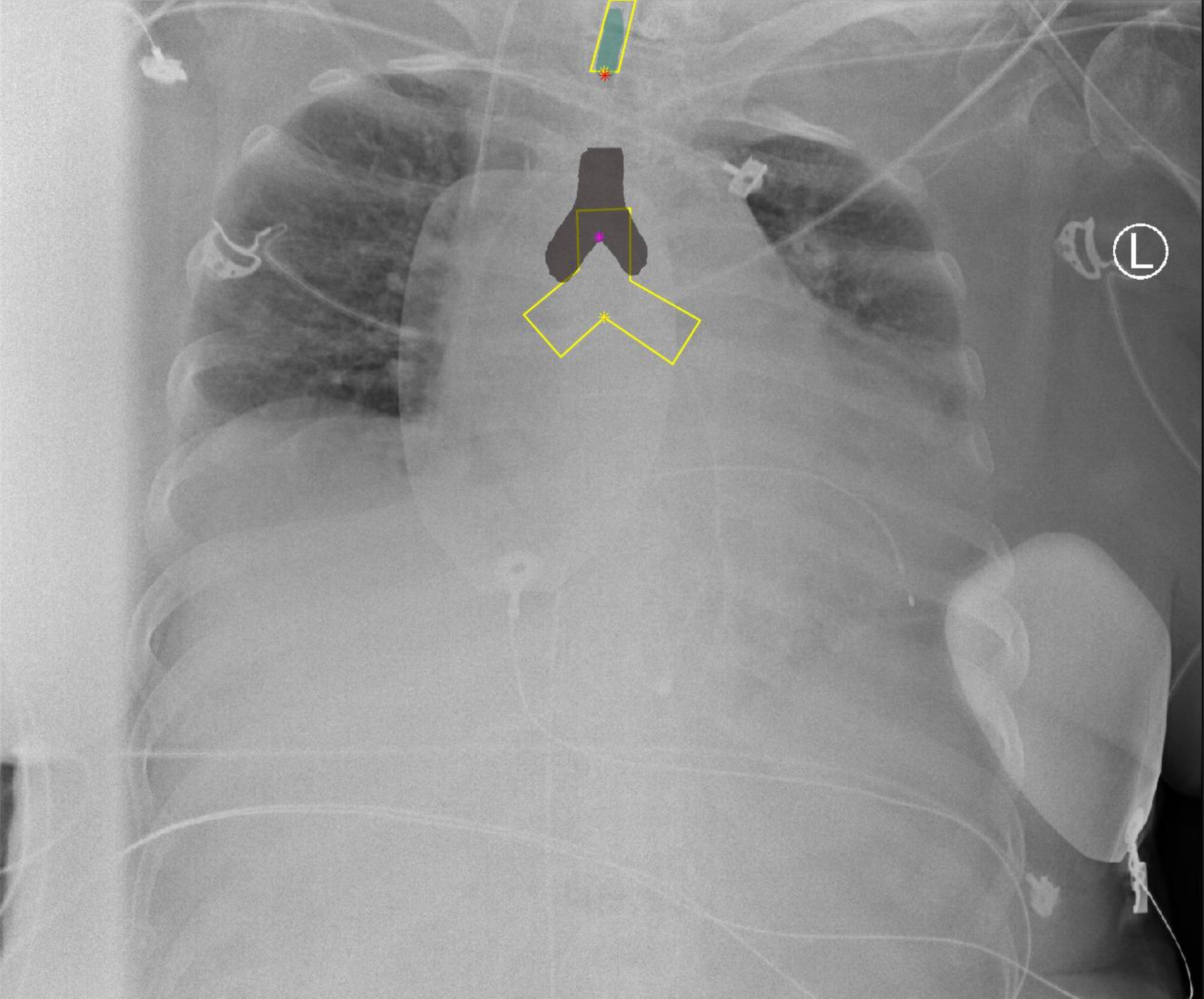} & \includegraphics[width=2 in]{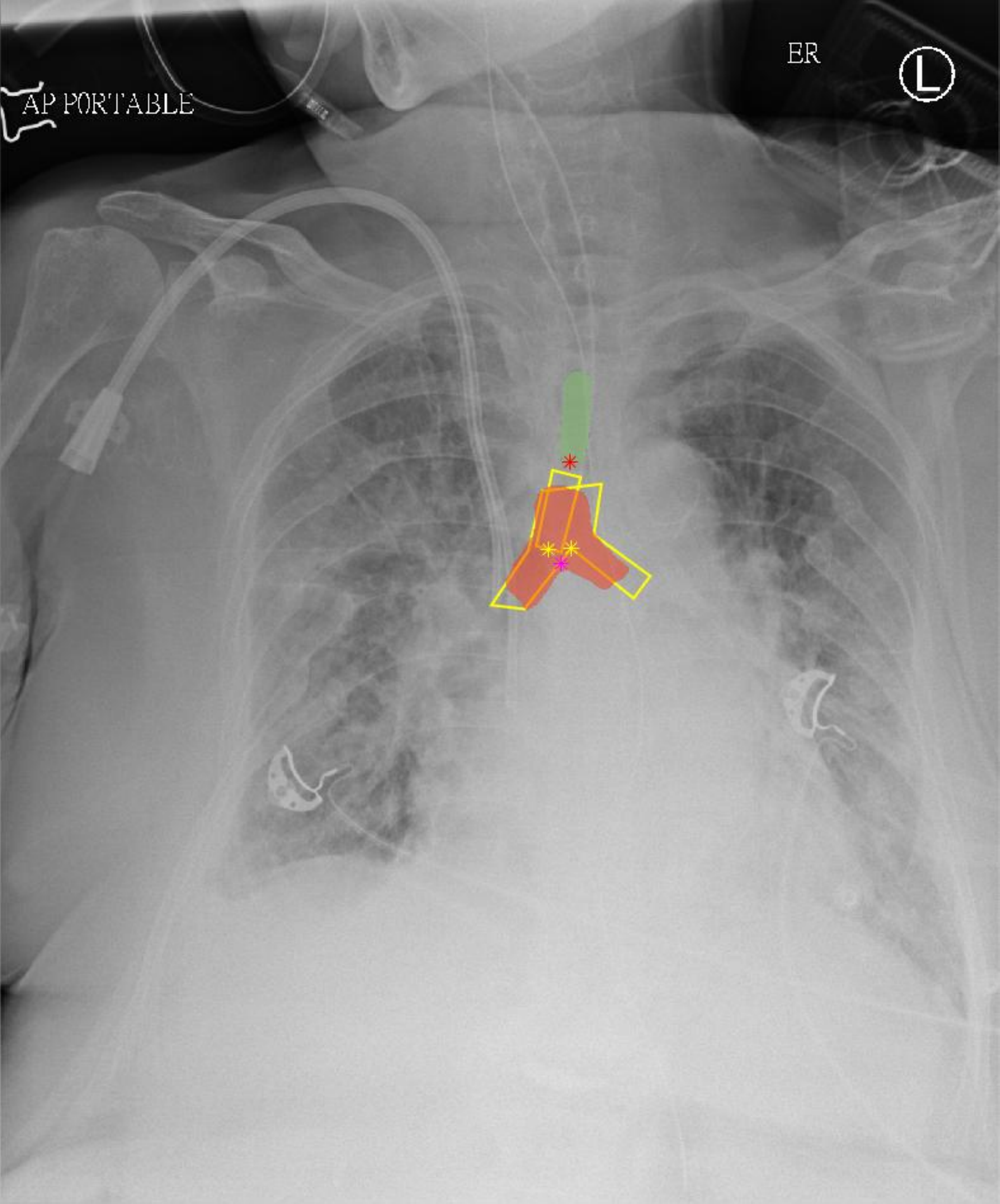} & \includegraphics[width=2 in]{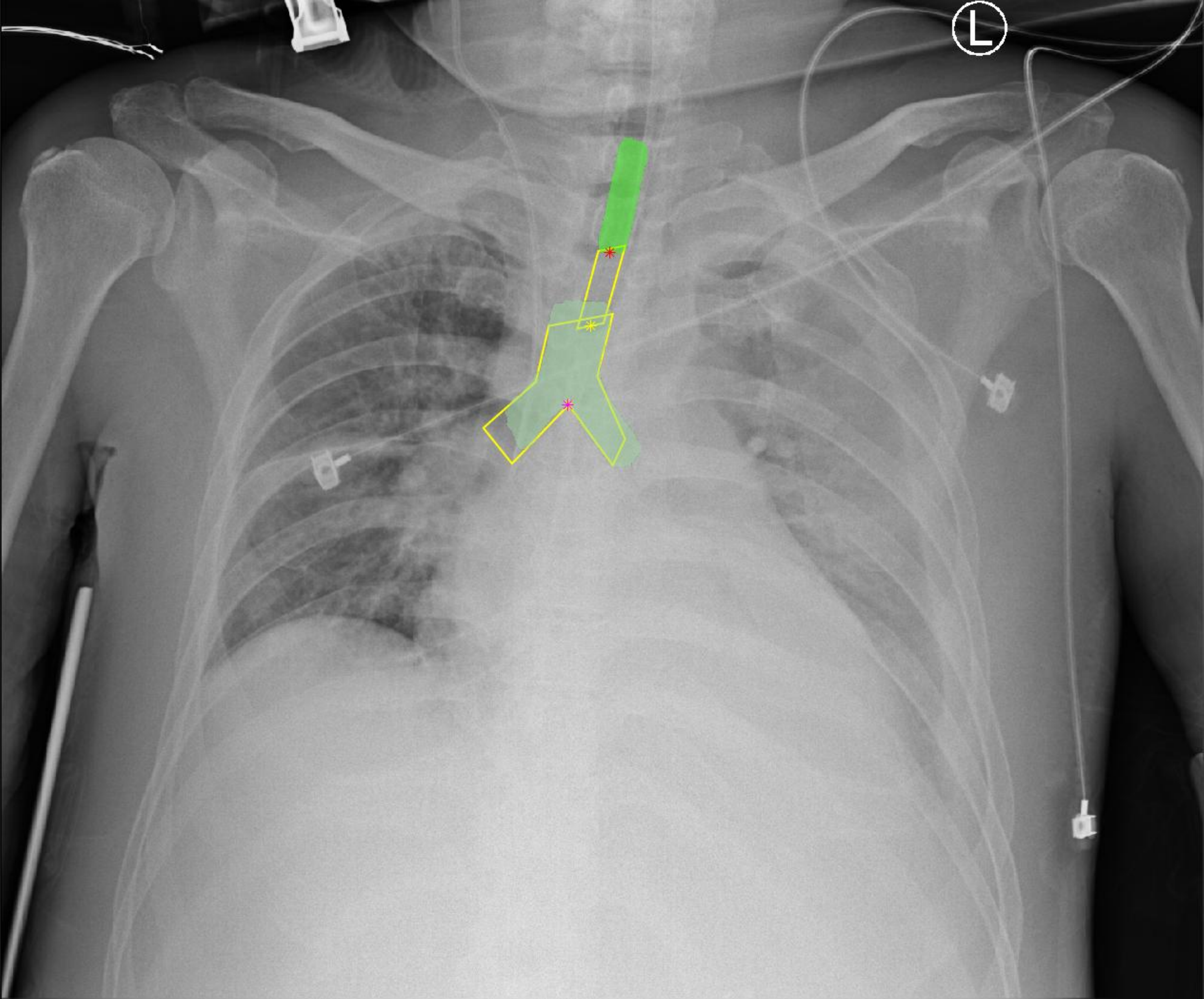} \\ \hline
\end{tabular}
\label{tab:ratio10}
\end{table*}

\section{Discussion and Conclusion}\label{sec:Conclusion}
For most critically ill patients, endotracheal intubation with mechanical ventilation is crucial to maintain their lives. Malposition of the ETT can cause serious harm to the patients if not detected in a timely manner. Clinicians need to verify proper ETT positioning by measuring the distance between the ETT tip and the carina on a portable supine CXR. If malposition is detected, clinicians need to adjust the ETT position according to measured distance. However, it is not easy to clearly identify the positions of the ETT and the carina on portable supine CXRs due to the low image contrast and abundant noise at the region of interest, especially for clinicians with less experience. An AI system to reliably identify ETT and carina positions on portable supine CXRs may lead to rapid and accurate identification of improper ETT position and reduce the risk of critically ill patients.

This paper proposed a feature extraction method with Mask R-CNN to predict feature points of tube and tracheal bifurcation in an image. In our experiments, our result exceeded 96\% in terms of recall and precision. The object error was less than $4.7751\pm 5.3420$ mm, and the ETT-carina distance error was less than $5.5432\pm 6.3100$ mm. Moreover, the proposed system labeled the feature point of the ETT tip and that of the carina, which can assist intensivists to monitor ICU patients' condition. According to the Pearson correlation coefficient, we have a strong correlation between the board-certified intensivists and our result in terms of ETT-carina distance. The external validation shows that the proposed method is a high-robustness system.

Although the proposed AI system can overcome problems from a portable CXR with lower contrast and higher noise, the system cannot operate normally in CXRs with extremely poor quality due to the improper operation of the X-ray machine. Therefore, future work is to suggest a clinical guideline and to obtain feedback from medical staff.

\section*{Acknowledgment}
The work was supported by the Ministry of Science and Technology, Taiwan, R.O.C. under Grants no. MOST 110-2634-F-006-012 - and by Higher Education Sprout Project, Ministry of Education to the Headquarters of University Advancement at National Cheng Kung University (NCKU).

\end{document}